\documentclass{article}

\usepackage{multirow} 
\usepackage{enumitem}

\newcommand{\eg}{\textit{e}.\textit{g}.,\ }
\newcommand{\alf}{ALFWorld}  
\newcommand{\sci}{SciWorld}  
\newcommand{\app}{AppWorld}  

\usepackage{array}      % provides \newcolumntype
\usepackage[table]{xcolor} % provides \columncolor (table option enables it cleanly in tabular)

\definecolor{lightblue}{RGB}{130,169,217}
\definecolor{green}{RGB}{29,177,0}

\definecolor{Gray}{gray}{0.92}
\definecolor{racing-green}{rgb}{0.0, 0.8, 0.6}
\definecolor{awesome-red}{rgb}{1.0, 0.13, 0.32}
\definecolor{LightCyan}{rgb}{0.88,1,1}
\definecolor{darkgreen}{RGB}{0,150,0}
\definecolor{Ground}{RGB}{255,184,55}
\definecolor{Dirt}{RGB}{191,169,115}
\definecolor{Pink}{RGB}{226,184,176}
\definecolor{Violet}{RGB}{163,148,170}
\definecolor{darkred}{RGB}{150,0,0} % 
\definecolor{bluelight}{RGB}{240,240,255}
\definecolor{greenight}{RGB}{240,255,240}

\usepackage{multirow}
\usepackage{siunitx}
\usepackage{adjustbox} % 比 resizebox 更稳：不会强行放大到超清晰度以外
\usepackage{makecell}

\newcommand{\prompttext}[1]{\ttfamily #1}           % Static Prompt Text
\newcommand{\promptvar}[1]{\textsc{#1}}             % Dynamic Variables
\newcommand{\metainstruct}[1]{\normalfont\itshape\color{darkgray} #1} % Meta/Logic
\usepackage{microtype}
\usepackage{graphicx}
\usepackage{subcaption}
\usepackage{booktabs} % for professional tables

\usepackage{hyperref}

\usepackage[preprint]{icml2026}

\usepackage{amsmath}
\usepackage{amssymb}
\usepackage{mathtools}
\usepackage{amsthm}

\usepackage[capitalize,noabbrev]{cleveref}

\theoremstyle{plain}

\theoremstyle{definition}

\theoremstyle{remark}

\usepackage[textsize=tiny]{todonotes}

\icmltitlerunning{Budget-Aware Agentic Routing via Boundary-Guided Training}

\begin{document}

\twocolumn[
  \icmltitle{Budget-Aware Agentic Routing via Boundary-Guided Training}



  \icmlsetsymbol{equal}{*}

  \begin{icmlauthorlist}
    \icmlauthor{Caiqi Zhang}{cambridge}
    \icmlauthor{Menglin Xia}{M365}
    \icmlauthor{Xuchao Zhang}{M365}
    \icmlauthor{Daniel Madrigal}{M365}
    \icmlauthor{Ankur Mallick}{M365}
    \icmlauthor{Samuel Kessler}{M365}
    \icmlauthor{Victor R{\"u}hle}{M365}
    \icmlauthor{Saravan Rajmohan}{M365}
  \end{icmlauthorlist}

  \icmlaffiliation{cambridge}{University of Cambridge}
  \icmlaffiliation{M365}{M365 Research, Microsoft}

  \icmlcorrespondingauthor{Menglin Xia}{mollyxia@microsoft.com}

  \icmlkeywords{Machine Learning, ICML}

  \vskip 0.3in
]



\printAffiliationsAndNotice{}  % no special notice (required even if empty)

% Notice that we need to explain:
% 1. why we need soft and hard budget;
% 2. why we call this boundary guided; 

\begin{abstract}
As large language models (LLMs) evolve into autonomous agents that execute long-horizon workflows, invoking a high-capability model at every step becomes economically unsustainable. 
While model routing is effective for single-turn queries, agentic routing is a sequential, path-dependent problem: early mistakes compound, feedback is often at the end of the episode, and deployments often demand strict per-task spending limits. 
We propose \textit{Budget-Aware Agentic Routing}, which selects between a cheap and an expensive model at each step to optimize the cost--success frontier and to operate under strict per-task budgets.
We propose Boundary-Guided Training, which leverages two boundary policies (always-small vs.\ always-large) to build a difficulty taxonomy and to anchor learning under sparse rewards. 
Our approach warms start with boundary-guided SFT data synthesis via stratified sampling of cost-efficient trajectories, then applies \textit{Boundary-Guided Policy Optimization} (BoPO), combining boundary-relative rewards with a reference-guided advantage to avoid degenerate cheap-failure solutions.
Experiment results show that our method improves the efficiency frontier, matching strong routing baselines at substantially lower cost while demonstrating generalization to strict inference-time budget constraints. Overall, our work establishes a foundational framework for agentic routing, shifting the paradigm from static model selection to dynamic, budget-aware sequential decision-making.

\end{abstract}

\section{Introduction}

Large language models (LLMs) are increasingly deployed not just to answer single questions, but to \emph{act} as agents that plan, call tools, write code, and revise intermediate decisions over long interaction trajectories \citep{ALFWorld20, wang-etal-2022-scienceworld, trivedi-etal-2024-appworld}. In these agentic workflows, a single task can require numerous steps, and each step may involve an expensive model call. This creates a \textbf{core deployment tension}: using a strong model everywhere is often prohibitively costly, while using a cheap model everywhere can fail on the few critical steps where advanced reasoning is necessary. The practical goal is therefore \emph{step-wise model selection}: allocate a small number of high-capability calls to the steps that matter most, while routing the remaining steps to cheaper models.

Most existing routing work, however, is optimized for \emph{single-turn} decisions (choose one model per query) \citep{chen2024frugalgpt, dingbest, ding2024hybrid} or for short-horizon orchestration \citep{zhang2025routerr, qian2025xrouter}. Agentic routing is fundamentally different. First, routing decisions are \emph{path-dependent}: early mistakes affect later states, so the value of a model choice depends on the evolving trajectory. Second, agentic environments typically provide only sparse terminal feedback, making credit assignment over long horizons extremely challenging. Third, real deployments impose diverse budget considerations, ranging from flexible cost--performance trade-offs to strict per-task spending caps. Thus, we are in need of a new formulation that treats model routing as a sequential decision process rather than a one-off / greedy classification in agentic tasks.

In this paper we study \textbf{Budget-Aware Agentic Routing (BAAR)}: given a multi-step trajectory, a router repeatedly chooses between a cost-efficient model and a highly capable but expensive model to maximize task success under compute costs. We formalize two complementary paradigms. \emph{Soft-budget routing} optimizes an efficiency frontier by trading off success against cost, producing a Pareto curve of policies with different spending profiles. \emph{Hard-budget routing} captures strict constraints (e.g., a maximum budget per task), where the router must manage a limited resource throughout the trajectory to achieve higher task performance. 

To make BAAR trainable under sparse rewards, we introduce \textbf{Boundary-Guided Training}, a two-stage pipeline that anchors learning to meaningful performance boundaries. % instead of relying on fragile exploration. 
The key idea is to construct \emph{reference boundaries} using two extreme policies (always using the small model and always using the large model) and use these boundaries to shape both offline data synthesis and online policy updates. Concretely, we first profile tasks into an intrinsic difficulty taxonomy (Easy/Hard/Intractable). We then propose \textit{Boundary-Guided Supervised Fine-Tuning (BoSFT)}: for Easy and Intractable tasks, the cheap boundary provides an unambiguous label (solve cheaply when possible; otherwise fail cheaply), while for Hard tasks we synthesize cost-effective proxies that approximate the optimal efficient paths vis stratified sampling.
Starting from this warm start, we introduce \textit{Boundary-Guided Policy Optimization (BoPO)}, an online reinforcement learning method that prevents the common ``always-small'' collapse under sparse terminal rewards. BoPO uses (i) a boundary-relative reward that calibrates cost penalties against task difficulty so the router spends aggressively only when it is likely to change outcomes, and (ii) a reference-guided advantage that compares sampled trajectories not only to their group baseline but also to the offline critical-path reference. Finally, to support strict deployment limits, we enforce strict budgets at inference time with \textit{Budget-Constrained Decoding}, which prunes actions that would violate the remaining budget.

We evaluate BAAR on three long-horizon interactive agent benchmarks spanning scientific discovery, embodied instruction following, and tool-using coding. Across all environments, BoPO consistently improves the cost--success efficiency frontier, matching the performance of always using the large model at a fraction of the cost. Under strict hard-budget limits, BoPO transfers well without hard-budget-specific training, achieving strong success rates; however, we also observe a limitation: policies trained with a fixed soft-budget trade-off do not fully adapt to different hard caps, motivating routers that explicitly condition on remaining budget in the future work.

\textbf{Contributions.}
We summarize our main contributions as follows:
(i) We formalize \emph{Budget-Aware Agentic Routing} as a long-horizon sequential decision problem and distinguish soft-budget and hard-budget deployment paradigms.
(ii) We propose Boundary-Guided Training, combining a difficulty taxonomy with stratified-sampling SFT data synthesis (BoSFT) and a boundary-anchored policy optimization method (BoPO) to stabilize learning under sparse feedback. 
(iii) We demonstrate consistent gains on three benchmarks, pushing the efficiency frontier by conserving budget on easy tasks and funding complex reasoning on hard ones. 
\section{Related Work}

\textbf{Single-Turn Routing.}
Most LLM routing work treats routing as a \emph{single-query} decision: given an input, choose one model that best balances quality and cost. Early foundations like FrugalGPT proposed cascading strategies to sequentially call models of increasing cost only when necessary \cite{chen2024frugalgpt}. Subsequent work has refined this via preference data learning \cite{ong2025routellm}, unified routing-cascading frameworks \cite{dekoninck2025a}, and quality-aware routing that dynamically mixes models to optimize single-turn response quality \cite{ding2024hybrid, wang-etal-2025-mixllm}. As routing systems mature, standardized benchmarks like RouterBench \cite{hu2024routerbench} and RouterEval \cite{huang-etal-2025-routereval} also emerge, which systematically assess router performance across diverse domains. 
However, the above mentioned works largely optimize \textbf{per-query} assignments; in contrast, our setting goes beyond single-turn selection to \emph{agentic routing}, where routing decisions are made repeatedly across a multi-step trajectory.   

\textbf{Agentic and Multi-Step Routing.}
The transition from answering questions to solving tasks has spurred interest in routing for autonomous systems. MasRouter addresses this in multi-agent systems by routing sub-tasks to specialized agent roles \cite{yue-etal-2025-masrouter}. More closely related to our work are recent RL-based orchestration systems like xRouter \cite{qian2025xrouter} and Router-R1 \cite{zhang2025routerr}, which train routers to manage multi-round interactions and aggregation. 
Our work diverges from these approaches by addressing much longer trajectories (averaging 20 steps) and targeting the \textit{efficiency frontier} of step-wise model selection.
Specifically, we optimize the allocation of expensive compute across intermediate steps to minimize costs while maintaining performance.

\textbf{Budget-Constrained Routing.}
Practical routing must satisfy hard constraints such as cost~\cite{chen2024frugalgpt, ong2025routellm}, latency~\cite{wang-etal-2025-mixllm, lakha2025faster}, and throughput/capacity~\cite{omni}, motivating constrained-optimization formulations.
Recent approaches enforce budgets via adaptive routing policies \cite{panda-etal-2025-adaptive} or test-time compute optimization \cite{dingbest}, and related work considers joint cost--latency constraints \cite{lakha2025faster}.
Frameworks such as OmniRouter \citep{omni} further cast routing as a \emph{global} constrained optimization problem over a query distribution rather than a sequence of per-query greedy decisions, while CARROT \citep{somerstep2025carrot} provides a lightweight cost--quality trade-off router with theoretical guarantees.
However, these methods typically allocate resources \emph{across independent queries}; in contrast, we allocate a limited number of high-capability calls \emph{within a single long-horizon agent run}, where early routing mistakes propagate and later steps depend on earlier tool/model choices.

\section{Budget-Aware Agentic Routing (BAAR)}

We formulate dynamic model routing for agentic workflows as a sequential decision process over a distribution of tasks $x \in \mathcal{D}$. Each task $x$ initiates an interaction trajectory that may span up to $T$ steps (typically $T > 20$ for complex reasoning workflows).

\textbf{State and Action Space.} At time step $t$, the router operates on the interaction history $s_t$—comprising the task input, past actions, and tool outputs, etc. In our Partially Observable Markov Decision Process (POMDP) formulation, $s_t$ acts as the effective state representation for the policy $\pi_\theta(a_t \mid s_t)$. Based on this state, the router selects an action $a_t \in \mathcal{A} = \{\mathcal{M}_{\text{small}}, \mathcal{M}_{\text{large}}\}$, deciding between a cost-efficient model and a highly capable but expensive one.

\textbf{Routing for Efficiency.} Unlike existing works that prioritize maximizing accuracy with a pool of LLMs \citep{qian2025xrouter, zhang2025routerr}, our objective in BAAR is to optimize the \textit{efficiency frontier}. In this paper, we focus on the binary choice in $\mathcal{A}$ to isolate the fundamental economic trade-off in agentic deployment: identifying the minimal critical steps of expensive compute required to solve a task. While our framework is extensible to multiple models, this setting captures the core challenge of minimizing cost without compromising success on complex reasoning tasks.

\textbf{Dynamics.} Executing action $a_t$ generates the next observation and transitions the system according to $s_{t+1} \sim P(\cdot \mid s_t, a_t)$, incurring a scalar cost $c(a_t) > 0$. A complete trajectory is denoted as $\tau = \{(s_t, a_t)\}_{t=0}^{|\tau|-1}$, which terminates upon task completion or reaching the horizon $T$. 

To bridge the gap between flexible training objectives and strict real-world deployment constraints, we formalize two distinct routing paradigms.

\subsection{Soft-Budget Routing (Efficiency Frontier)}
\label{sec:soft_routing}

Intuitively, this paradigm addresses the trade-off between performance and resource consumption. It seeks to answer: \textit{How can we maximize task performance if we incur a soft penalty for every unit of compute consumed?}

We formalize this as an unconstrained optimization problem where the goal is to maximize a scalar reward that linearly combines success and cost. The policy optimizes the following objective parameterized by a Lagrange multiplier $\lambda \ge 0$:
\begin{equation}
J_{\text{soft}}(\theta) =
\mathbb{E}_{\tau \sim \pi_\theta}\!\left[
\mathbb{I}(\text{success}(\tau)) - \lambda \sum_{t=0}^{|\tau|-1} c(a_t)
\right],
\label{eq:soft}
\end{equation}
where $\mathbb{I}(\text{success}(\tau)) \in \{0,1\}$ is a sparse terminal indicator function that is $1$ if the task is solved and $0$ otherwise. Here, $\lambda$ controls the trade-off intensity: varying $\lambda$ traces an efficiency frontier, encouraging the router to utilize expensive compute only when the marginal gain in success probability outweighs the cost.

\subsection{Hard-Budget Routing (Strict Constraints)}
\label{sec:hard_routing}

While soft-budget routing provides a flexible cost--accuracy trade-off, real-world deployments often impose \emph{strict} spending limits. This paradigm captures that setting:
\textit{given a fixed, non-negotiable budget (e.g., \$0.50 per task), how can we maximize the success rate?}

We formalize this setting as a Constrained Markov Decision Process (CMDP). Unlike the soft-budget case, the objective here is to maximize success subject to a strict upper bound on the total trajectory cost:
\begin{equation}
\max_{\pi}\ \mathbb{E}_{\tau \sim \pi}\!\left[\mathbb{I}(\text{success}(\tau))\right]
\quad \text{s.t.} \quad
\sum_{t=0}^{|\tau|-1} c(a_t) \le B_{\max},
\label{eq:hard}
\end{equation}
where $B_{\max}$ represents the maximum allowable budget per task (e.g., API cost limit, numbers of calling large models).

\textbf{Optimization Challenge.} The central challenge in both paradigms is \textit{sparse credit assignment}: trajectory success depends on a sequence of heterogeneous routing choices, while intermediate feedback is unavailable. Furthermore, the CMDP formulation in Eq.~\eqref{eq:hard} introduces the difficulty of dynamically managing a budget resource over a long horizon under uncertainty. 

\section{Methodology}

We now detail our training pipeline, which consists of two phases: Boundary-Guided Supervised Fine-Tuning (BoSFT) and Boundary-Guided Policy Optimization (BoPO). Our approach introduces three key innovations to address the budget-constrained routing problem: 1) a granular \textbf{difficulty taxonomy} to characterize task hardness; 2) a \textbf{Stratified Sampling} mechanism to synthesize efficient routing data; and 3) the \textbf{BoPO} algorithm for optimization. 

\subsection{Task Difficulty Taxonomy}
\label{sec:taxonomy}

Prior to training, we characterize the intrinsic difficulty of tasks in the distribution $\mathcal{D}$. This profiling filters noise for SFT data synthesis and establishes performance anchors for the reward functions used in BoPO.

We probe each task $x$ by executing two deterministic policies for all steps: $\pi_{\text{small}}$ (exclusively routing to $\mathcal{M}_{\text{small}}$) and $\pi_{\text{large}}$ (exclusively utilizing $\mathcal{M}_{\text{large}}$). To mitigate stochasticity, we execute both baselines $K=5$ times for every query. Based on the aggregate success rates, we partition the task distribution into three disjoint sets:

\begin{enumerate}[leftmargin=1.2em, itemsep=0em, labelsep=0em, topsep=0em]
    \item \textbf{Easy ($\mathcal{D}_{\text{easy}}$):} Tasks where $\pi_{\text{small}}$ succeeds in at least 4 out of 5 runs (Pass@5 $\ge 0.8$). For these queries, the cost-efficient model is sufficient; routing to the larger model is computationally wasteful.
    \item \textbf{Hard ($\mathcal{D}_{\text{hard}}$):} Tasks where $\pi_{\text{small}}$ is unreliable ($0 \leq $ Pass@5 $< 0.8$) but $\pi_{\text{large}}$ could succeed. These tasks require the reasoning capabilities of $\mathcal{M}_{\text{large}}$ to be solved correctly.
    \item \textbf{Intractable ($\mathcal{D}_{\text{intractable}}$):} Tasks where even $\pi_{\text{large}}$ fails across all trials. Since neither model succeeds, the optimal strategy is to ``fail cheaply" using $\mathcal{M}_{\text{small}}$ rather than incurring high costs for a guaranteed failure.
\end{enumerate}

\subsection{Boundary-Guided Supervised Fine-Tuning (BoSFT)}
\label{sec:sft}

To circumvent the cold-start problem inherent in RL, we initialize the router via SFT. However, standard behavioral cloning is insufficient because optimal routing decisions are often latent. To address this, we propose \textit{Boundary-Guided SFT}, a data synthesis method that explicitly approximates the optimal decision boundary.

\textbf{Motivation.}
For tasks in $\mathcal{D}_{\text{easy}}$ and $\mathcal{D}_{\text{intractable}}$, constructing target trajectories $\tau^*$ is trivial. We assign the trajectory generated by $\pi_{\text{small}}$ as the expert label. For $\mathcal{D}_{\text{easy}}$, this teaches the router that the cost-efficient model is sufficient; for $\mathcal{D}_{\text{intractable}}$, it instills a ``fail-cheap'' mechanism: if a task is beyond the system's capabilities, the agent should minimize cost rather than blindly scaling up.
However, this approach fails for $\mathcal{D}_{\text{hard}}$. Simply cloning the $\pi_{\text{large}}$ trajectory is suboptimal because it likely contains redundant expensive calls. Our objective is to discover a \textit{cost-effective proxy} that approximates the optimal path by eliminating obvious computational waste.

\textbf{Stratified Sampling for Hard Tasks.} 
To approximate this minimal cost path, we employ \textit{Stratified Sampling}. We generate $N$ (\eg 10, 20) independent trajectories where the router samples model choices stochastically according to a probability sweep $p_k$:
\begin{equation}
    P(a_t = \mathcal{M}_{\text{large}}) = p_k = \frac{k}{N}, \quad k \in \{1, \dots, N\}.
\end{equation}
By varying $p_k$ from $0$ to $1$, we explore the full spectrum of routing behaviors, from mostly small to mostly large usage. From the set of resulting successful trajectories $\mathcal{T}_{\text{succ}}$, we select the expert trajectory $\tau^*$ that minimizes total cost:
\begin{equation}
    \tau^* = \operatorname*{argmin}_{\tau \in \mathcal{T}_{\text{succ}}} \sum_{t=0}^{|\tau|-1} c(a_t).
\end{equation}
This process identifies a proxy for the optimal policy: while $\tau^*$ significantly reduces cost compared to the static $\pi_{\text{large}}$, it remains an upper bound on the true optimal cost due to sampling limitations. We accept this proxy for initialization, leaving further optimization to the RL phase in Section~\ref{sec:bopo}.

\subsection{Boundary-Guided Policy Optimization (BoPO)}
\label{sec:bopo}

While BoSFT provides a warm start, the resulting policy remains limited by the static nature of the offline dataset. The router cannot adapt to run-time stochasticity or discover novel, lower-cost routing sequences that were absent from the stratified sampling phase. To address this, we introduce \textit{Boundary-Guided Policy Optimization (BoPO)}, an online reinforcement learning framework designed explicitly for the sparse-reward, cost-constrained nature of agentic routing.

\textbf{Overview.} We build BoPO on {Group Relative Policy Optimization (GRPO)} because it avoids a learned value function, which is beneficial in our setting where the effective state $s_t$ includes long interaction histories ($T \approx 20$). However, vanilla GRPO compares trajectories only within a sampled group, which is fragile under sparse terminal feedback: if an entire group fails, the algorithm may still reinforce the cheapest failure mode and drift toward a degenerate ``always-small'' policy. BoPO resolves this by introducing \emph{boundary-guided reference anchors} derived from the two extreme policies $\pi_{\text{small}}$ (exclusively routing to $\mathcal{M}_{\text{small}}$) and $\pi_{\text{large}}$ (exclusively utilizing $\mathcal{M}_{\text{large}}$). These anchors keep the policy updates aligned with a meaningful notion of progress: improving success beyond the $\pi_{\text{small}}$ boundary, while paying only for compute that is necessary.

\textbf{Boundary-Relative Reward.}
A central challenge in routing is unifying the reward signal across heterogeneous tasks where computational costs vary by orders of magnitude. We propose a \textit{Boundary-Relative} objective that dynamically calibrates the reward magnitude based on the task classification derived in Section~\ref{sec:taxonomy}.
For a generated trajectory $\tau$ given context $x$, the reward function is:
\begin{equation}
    R(\tau, x) = (r_{\text{success}} + r_{\text{hard}}) - \lambda \cdot \mathcal{C}_{\text{norm}}(\tau, x).
    \label{eq:reward_final}
\end{equation}
Here, $r_{\text{success}}$ represents a base utility for completing task $x$ correctly, while $r_{\text{hard}}$ acts as an additional bonus (assigned a positive value only when $x \in \mathcal{D}_{\text{hard}}$ and zero otherwise). We introduce these distinct terms to decouple the value of basic correctness from the value of complex reasoning, preventing the ``laziness" often induced by uniform rewards.

This structure also enforces a strict efficiency frontier (see Appendix~\ref{app:reward_analysis} for an analysis of the reward landscape):
(1) \textit{For Easy Tasks ($r_{\text{hard}}=0$):} The total potential reward is capped at the small base $r_{\text{success}}$. This tight margin forces the optimizer to minimize the cost term $\mathcal{C}_{\text{norm}}$ to secure a positive return.
(2) \textit{For Hard Tasks ($r_{\text{hard}}>0$):} The bonus of solving $r_{\text{hard}}$ compensates the cost penalty, strictly prioritizing task success over budget conservation when advanced reasoning is required.

To ensure the cost penalty scales appropriately across tasks with varying lengths, we define the \textbf{Normalized Cost} $\mathcal{C}_{\text{norm}}$ as the fraction of the ``risk budget'' consumed relative to the static baselines. 
% Let $C_{\min}(x)$ and $C_{\max}(x)$ denote the minimum and maximum median costs among the extreme policies:
Let $C_{\min}(x)$ and $C_{\max}(x)$ denote the lower and upper boundary costs, defined as the minimum and maximum of $\{C_{\text{small}}(x),\, C_{\text{large}}(x)\}$, respectively. The normalized cost is then defined as:
\begin{equation}
    \mathcal{C}_{\text{norm}}(\tau, x) = \text{clip}\left( \frac{C(\tau) - C_{\min}(x)}{C_{\max}(x) - C_{\min}(x) + \epsilon}, \; 0, \; 1 \right).
\end{equation}
Here, $C(\tau) = \sum_{t} c(a_t)$ is the trajectory cost. This formulation renders the penalty scale-invariant and robust to cost inversions (e.g., when $\pi_{\text{small}}$ is inefficient due to long failure trajectories). The clipping ensures that matching the efficiency of the best static baseline yields zero penalty, while ``runaway'' trajectories saturate at a maximum penalty of $\lambda$, preserving gradient stability.

\textbf{Reference-Guided Advantage.}
Standard GRPO computes the advantage $A(\tau_g)$ by comparing a trajectory's reward solely against the group mean $\mu_{\text{group}}$. In BoPO, we enforce a stricter standard by incorporating the offline expert trajectory $\tau^*$ from the SFT phase as a strict reference. For a sampled trajectory $\tau_g$ within a group of size $G$, the advantage is computed as:
\begin{equation}
    A(\tau_g) = \frac{R(\tau_g, x) - \max\left(\mu_{\text{group}}, R(\tau^*, x)\right)}{\sigma_{\text{group}} + \epsilon}.
\end{equation}
By referencing the maximum of the group mean and the expert reward $R(\tau^*, x)$, we penalize any policy update that degrades performance below the known ``critical path'' discovered during supervised training. This mechanism effectively prevents catastrophic forgetting of the optimal boundaries learned in SFT, while still allowing the policy to accrue positive advantage if and only if it discovers a novel routing strategy that is strictly superior to the reference.

\textbf{Objective Function.}
We optimize the policy $\pi_\theta$ using the GRPO objective augmented with the reference-guided advantage and a token-level KL divergence penalty to maintain semantic coherence:
\begin{equation}
\begin{split}
    J_{\text{BoPO}}(\theta) = \mathbb{E}_{x \sim \mathcal{D}, \tau \sim \pi_{\text{old}}} \bigg[ \frac{1}{G} \sum_{g=1}^G \bigg( \frac{\pi_\theta(\tau_g)}{\pi_{\text{old}}(\tau_g)} A(\tau_g) \\
    - \beta_{\text{KL}} D_{\text{KL}}\left(\pi_\theta || \pi_{\text{ref}}\right) \bigg) \bigg]
\end{split}
\end{equation}
where $\pi_{\text{ref}}$ is the SFT model. Through this formulation, BoPO enables the router to safely explore the efficiency frontier, learning to {conserve budget on trivial or intractable queries} (where scale offers no advantage) while {spending wisely on reasoning-intensive tasks} (where the larger model is critical for success).

\subsection{Hard Budget Routing via Constrained Decoding}
\label{sec:bcd}

While our primary training objective optimizes the efficiency frontier (Soft-Budget), real-world deployments often mandate strict cost ceilings, as formalized in the Constrained MDP (Eq.~\eqref{eq:hard}). Directly solving this constrained optimization problem via reinforcement learning is inherently unstable, as the global budget constraint induces a discontinuous objective and high-variance gradients (see Appendix~\ref{app:direct_hard_training} for a theoretical analysis).

To bridge this gap, we propose \textbf{Budget-Constrained Decoding (BCD)}. Instead of encoding hard constraints during training, we enforce them mechanically at inference time by pruning the action space based on the remaining funds. We track the residual budget $b_t = B_{\max} - \sum_{i=0}^{t-1} c(a_i)$ and define the effective decoding policy $\pi_{\text{BCD}}$ as:
\begin{equation}
    \pi_{\text{BCD}}(a_t \mid s_t) = 
    \begin{cases} 
        \mathbb{I}(a_t = \mathcal{M}_{\text{small}}) & \text{if } c(\mathcal{M}_{\text{large}}) \geq b_t \\
        \pi_\theta(a_t \mid s_t) & \text{otherwise}
    \end{cases}
\end{equation}
where $\mathbb{I}(\cdot)$ is the indicator function. This intervention strictly prevents the router from selecting the large model if it triggers a budget overrun, forcing a fallback to the small model.

We implement this by embedding the current budget status $b_t$ directly into the system prompt at every turn (Appendix \ref{app:prompts}). We evaluate whether the policy, trained only on soft efficiency signals, can interpret the explicit budget state and throttle its own usage to achieve higher task performance.

% \section{Experiments}

% \textbf{Datasets:} We use Appworld \citep{trivedi-etal-2024-appworld}
% , Alfworld \citep{ALFWorld20}
% , and 
% Scienceworld \citep{wang-etal-2022-scienceworld} as datasets. For Alfworld and Sciworld we use the AgentGym framework \citep{xi2024agentgym}. [elaborate briefly what are these datasets and why they provide a comprehensive coverage]; More details regarding the dataset statistics and max allowed steps can be found in tab

% train test max steps
% Alfworld 2420 200 30
% Sciworld 2120 200 40
% Appworld 105 168 40

% \textbf{Models:} We use Qwen2.5-1.5B as router [why it is efficient] and Our method is not dependent on any two pair of small large models and we choose GPT4.1 / GPT4.1 mini as big and small models. 

% \textbf{Baselines: }
% -	Big / small only
% -	Random routing
% -	GPT-5 as router (zero shot)
% -	Rule-based router (route to big model when there is any failure)
% - confidence bases router 
% -	SFT baseline (all small model failed examples go to big)
% -	SFT task level only select small or big; 

% what else?

\begin{figure*}[ht]
    \centering
    \includegraphics[width=\linewidth]{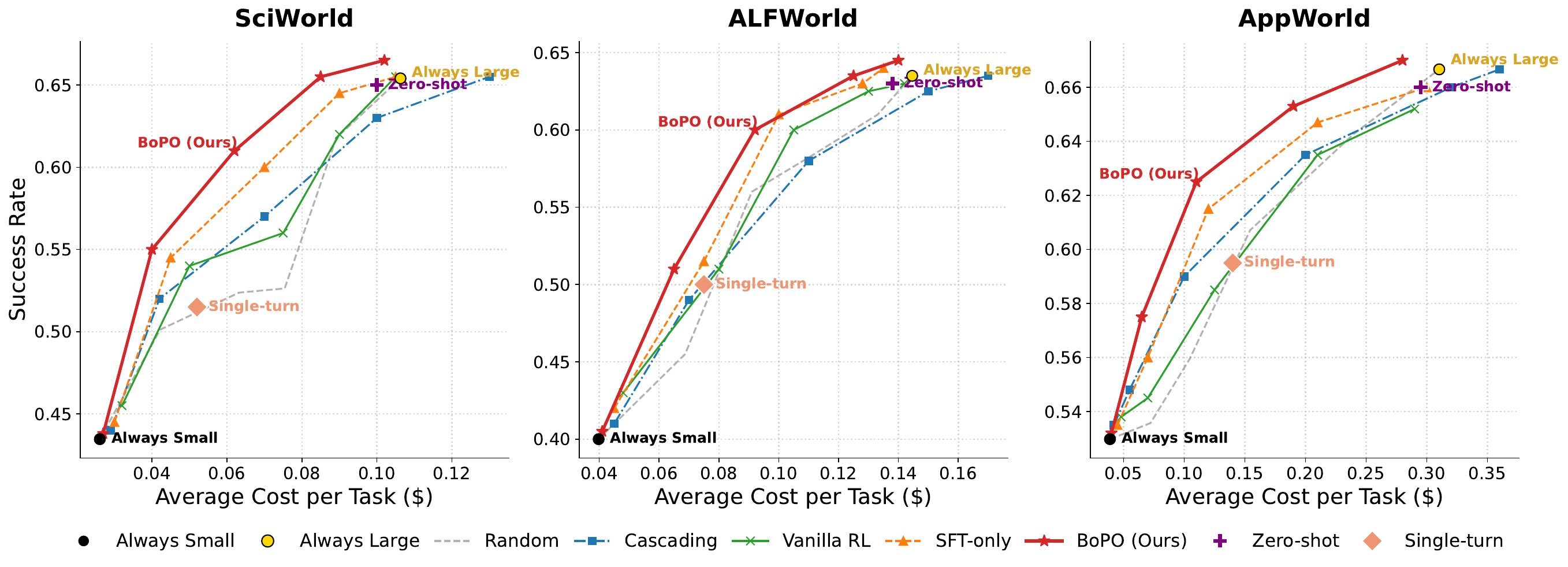}
    \caption{\textbf{Pareto efficiency frontiers across three agentic benchmarks.} We plot the Success Rate against the Average Cost per Task (\$) for SciWorld, ALFWorld, and AppWorld. \textbf{BoPO (Ours)} consistently pushes the efficiency frontier (top-left) compared to single-turn, cascading, and vanilla RL baselines, achieving comparable success to      ``Always Large" policies at a fraction of the cost.}    \label{fig:pareto_frontiers}
\end{figure*}

\section{Experiments}

\subsection{Setup}

\textbf{Environments and datasets.}
We evaluate on three complementary interactive agent benchmarks spanning  scientific reasoning, embodied decision-making, and interactive coding, under long-horizon interactions: \textbf{\sci} \citep{wang-etal-2022-scienceworld} focuses on elementary scientific discovery, where agents must design and execute experiments (e.g., testing conductivity), challenging the router's capacity for scientific reasoning and state tracking.
\textbf{\alf} \citep{ALFWorld20} aligns text-based instructions with embodied environments, requiring agents to plan and execute household tasks (e.g., ``clean the apple'') via navigation and object manipulation, testing the router's ability to handle spatial dependencies. 
\textbf{\app} \citep{trivedi-etal-2024-appworld} simulates a complex ecosystem of 9 interactive apps (e.g., Amazon, Gmail) via 457 APIs, requiring the agent to generate code, manage state, and handle high-fidelity API interactions; it serves as our primary testbed for tool-use and coding complexity. 
For \alf \ and \sci, we utilize the AgentGym framework \citep{xi2024agentgym} for standardized evaluation. More datasets details are in Table~\ref{tab:datasets}.

\textbf{Models and cost model.}
Our router policy $\pi_\theta(a_t \mid s_t)$ is instantiated with \textbf{Qwen2.5-1.5B} (instruction-tuned). 
% With only 1.5 billion parameters, it incurs negligible latency and computational overhead compared to the primary inference calls, ensuring that the router does not become a bottleneck in the agentic workflow.
Since our method is model-agnostic: we can pair any $(M_{\text{small}}, M_{\text{large}})$.
In particular, we use \textbf{GPT-4.1 mini} as $M_{\text{small}}$ and \textbf{GPT-4.1} as $M_{\text{large}}$.
We measure dollar cost using OpenAI list prices (per 1M tokens): GPT-4.1 is \$2.00/\$8.00 (in/output) and GPT-4.1 mini is \$0.40/\$1.60\footnote{\url{https://platform.openai.com/docs/pricing}}.
% implying a constant large-to-small price ratio of $\rho=5$.
Our agent is instantiated with ReAct-style prompting \citep{yao2022react}.

\textbf{Evaluation Protocols and Metrics.} 
Each task terminates upon environment success or reaching the maximum steps allowed. We report two primary metrics: 
(i) \textit{Success Rate (SR)}, and 
(ii) \textit{Average Cost}, defined as the expected monetary cost per task $\mathbb{E}_\tau[\sum_t c(a_t)]$.
In soft-budget setting, we vary the trade-off parameter $\lambda$ in Eq.~(1) or adjust the threshold to trace the Pareto efficiency frontier (SR vs. Cost) depending on different baselines. In hard-budget setting, we enforce strict per-task resource limits via Budget-Constrained Decoding (Eq.~(9)). In this setting, we define the budget $B_{\max}$ as the maximum allowable number of large model calls (denoted as $K$). 
All experiments are averaged over three random seeds.

\newcolumntype{B}{>{\columncolor{lightblue!5}}c}
\newcolumntype{G}{>{\columncolor{green!6}}c}
\newcolumntype{P}{>{\columncolor{purple!2.4}}c}
\newcolumntype{A}{>{\columncolor{bluelight!32}}c}
\newcolumntype{C}{>{\columncolor{greenight!32}}c}

\begin{table*}[ht]
\centering
\footnotesize
\caption{\textbf{Performance under strict hard-budget constraints.} We evaluate Success Rate (SR) and Large Model Usage ($Use\% = Avg. \#LargeCalls / K$) with a maximum budget of $K \in \{5, 10, 15\}$ large calls per task. BoPO generally outperforms baselines, though simple heuristics like First-Large remain competitive in strict settings. *Budget-Constrained Decoding (BCD) is applied to all settings except Always-Large. We list the Always-Large setting here only for reference and it can exceed the budget. }
\vspace{2mm}
\setlength{\tabcolsep}{2pt}
\renewcommand{\arraystretch}{1.35}

\begin{adjustbox}{max width=0.94\textwidth}
% \begin{tabular}{l
%                 SS SS SS
%                 SS SS SS
%                 SS SS SS}
\begin{tabular}{l
                BBBBBB
                GGGGGG
                PPPPPP}
\toprule
& \multicolumn{6}{c}{\textbf{ALFWorld}}
& \multicolumn{6}{c}{\textbf{SciWorld}}
& \multicolumn{6}{c}{\textbf{AppWorld}} \\
\cmidrule(lr){2-7}\cmidrule(lr){8-13}\cmidrule(lr){14-19}
& \multicolumn{2}{c}{\textbf{K=5}}  & \multicolumn{2}{c}{\textbf{K=10}} & \multicolumn{2}{c}{\textbf{K=15}}
& \multicolumn{2}{c}{\textbf{K=5}}  & \multicolumn{2}{c}{\textbf{K=10}} & \multicolumn{2}{c}{\textbf{K=15}}
& \multicolumn{2}{c}{\textbf{K=5}}  & \multicolumn{2}{c}{\textbf{K=10}} & \multicolumn{2}{c}{\textbf{K=15}} \\
\cmidrule(lr){2-3}\cmidrule(lr){4-5}\cmidrule(lr){6-7}
\cmidrule(lr){8-9}\cmidrule(lr){10-11}\cmidrule(lr){12-13}
\cmidrule(lr){14-15}\cmidrule(lr){16-17}\cmidrule(lr){18-19}
\rowcolor{white}
\textbf{Method}
& {\makecell[c]{SR}} & {\makecell[c]{Use\%}}
& {SR} & {Use\%}
& {SR} & {Use\%}
& {SR} & {Use\%}
& {SR} & {Use\%}
& {SR} & {Use\%}
& {SR} & {Use\%}
& {SR} & {Use\%}
& {SR} & {Use\%} \\
\midrule 
\textit{Always-Large*}
& 63.5 & 399
& 63.5 & 199
& 63.5 & 133
& 65.4 & 502
& 65.4 & 251
& 65.4 & 167
& 66.7 & 376
& 66.7 & 188
& 66.7 & 125 \\
\midrule
Always-Small
& 40.0 & 0
& 40.0 & 0
& 40.0 & 0
& 43.5 & 0
& 43.5 & 0
& 43.5 & 0
& 53.0 & 0
& 53.0 & 0
& 53.0 & 0 \\

First-Large
& 49.5 & 99
& 51.0 & 92
& 54.5 & 86
& 44.7 & 95
& 47.8 & 89
& 52.5 & 84
& 59.5 & 100
& 64.9 & 100
& 65.5 & 90 \\
\midrule

GPT-5
& 49.5 & 99
& 53.5 & 95
& 58.0 & 89

& 46.2 & 99
& 50.5 & 94
& 57.7 & 85

& 60.1 & 99
& 62.5 & 96
& 66.1 & 91 \\

Cascade
& 47.5 & 96
& 53.5 & 93
& 56.0 & 88

& 45.1 & 98
& 48.4 & 92
& 54.5 & 82

& 57.6 & 98
& 63.3 & 92
& 65.1 & 85 \\
Single-turn
& 50.0 & 98
& 52.5 & 92
& 57.5 & 84

& 44.0 & 99
& 47.4 & 93
& 53.8 & 86

& 58.8 & 99
& 63.3 & 94
& 64.9 & 82 \\
Vanilla RL
& 52.0 & 96
& 55.0 & 88
& 56.5 & 68

& 48.5 & 97
& 52.5 & 86
& 53.0 & 65

& 62.5 & 95
& 64.1 & 89
& 65.5 & 73 \\
\midrule

\textbf{BoPO} (Small $\lambda$)
& 48.0 & 99
& 54.5 & 98
& \textbf{60.0} & 92
& 45.5 & 98
& 51.0 & 96
& \textbf{59.5} & 88
& 59.0 & 97
& \textbf{65.0} & 96
& \textbf{66.5} & 88 \\
\textbf{BoPO} (Large $\lambda$)
& \textbf{54.2} & 96
& \textbf{56.6} & 75
& 55.5 & 55
& \textbf{50.4} & 92
& \textbf{55.1} & 72
& 54.5 & 45
& \textbf{64.2} & 94
& 64.8 & 75
& 65.0 & 60 \\

\bottomrule
\end{tabular}
\end{adjustbox}

\vspace{2pt}

\label{tab:hard_budget_sr_usepct_aligned}
\end{table*}

\textbf{Baselines.}
We compare against the representative routing baselines from different categories:
\textbf{(i) Always-small / Always-large} (route with $\pi_{\text{small}}$ or $\pi_{\text{large}}$ for all steps);
\textbf{(ii) Random routing} (route to $M_{\text{large}}$ with fixed probability $p$, swept to form a naive frontier);
\textbf{(iii) Zero-shot LLM router} (prompt a powerful model to output \textsc{SMALL}/\textsc{LARGE} given $(s_t,b_t)$, then execute the chosen agent model. While computationally expensive, this serves as a performance upper bound for zero-shot routers.);
\textbf{(iv) Cascading router} (FrugalGPT-style \citep{chen2024frugalgpt}: call $M_{\text{small}}$ first and escalate to $M_{\text{large}}$ only when a calibrated score falls below a tuned threshold);
\textbf{(v) Single-turn router} (RouteLLM-style \citep{ong2025routellm}: train a preference classifier on offline pairs $(s_t,\text{winner})$ and apply it greedily for step-wise decisions);
\textbf{(vi) SFT-only router} (our Boundary-Guided SFT policy without BoPO, testing the value of online optimization);
\textbf{(vii) Vanilla RL router} (optimize the standard cost-aware objective $\mathbb{I}(\text{success})-\lambda C$ using DAPO, \emph{without} boundary anchors or reference-guided advantage; similar to prior agentic routers that use DAPO/PPO \citep{qian2025xrouter, zhang2025routerr}. For the hard-budget setting, we additionally include a baseline \textbf{First-Large}, which uses the large model exclusively until reaching the budget $K$ to ensure the problem-solving process is on the right track, and then proceeds with small models. Note that for all baselines, the computational cost of the router itself is excluded from the reported metrics (more discussion in Appendix \ref{app:cost_router}).

More experiment details can be found in Appendix \ref{app:exp} and the prompts to the routers are in Appendix \ref{app:prompts}.

% baselines that are not suitable / not needed
% \item \textbf{Confidence-based Cascade:} route to $M_{\text{large}}$ when $M_{\text{small}}$'s self-assessed confidence is below a tuned threshold (e.g., via a calibrated verbal confidence head or a consistency check); otherwise use $M_{\text{small}}$.
% \item Task-level selector: maybe too weak, like a strawman
% \item Step-level supervised router: redundant
% \item Contextual Bandit Router (e.g., PILOT or MixLLM)
% PILOT: Srivatsa et al., Adaptive LLM Routing Under Budget Constraints (EMNLP 2025).
% MixLLM: Ding et al., MixLLM: Dynamic Routing in Mixed Large Language Models (NAACL 2025).

% ablations
% No taxonomy (train on all tasks)
% 	•	No stratified “boundary” SFT (standard SFT)
% 	•	BoPO without boundary reward shaping
% 	•	BoPO without reference-guided advantage
            % only stratified SFT
%           only RL without SFT

% 	•	Sensitivity: K (probe repeats), N (stratification granularity), λ, horizon length T, cost ratio between models
% (iii) \textbf{No stratified sampling:} replace Hard-task expert selection (Eq.~(4)) with the full $\pi_{\text{large}}$ trajectory (tests the importance of critical-path distillation).

\subsection{Results}
\label{subsec:results}

\textbf{BoPO dominates the efficiency frontier across all environments.}
We first analyze the trade-off between task success and computational cost in soft budget mode. As illustrated in the Pareto frontiers (Figure~\ref{fig:pareto_frontiers}), our method consistently yields higher success rates at lower costs compared to all baselines. For example, in ALFWorld, our method reaches the high-performance regime ($>63\%$) at a cost of $\$0.125$, effectively undercutting the other baselines. Regarding the Zero-shot GPT-5 router, we observe that its decisions are \textit{highly sensitive} to prompt design and prone to mode collapse; it typically defaults to a static strategy (exclusively selecting either the small or large model for an entire trajectory) rather than adapting dynamically. This lack of granular adaptivity underscores the necessity of a specialized, trained router.

\textbf{Boundary-guided training prevents policy collapse.} Figure \ref{fig:pareto_frontiers} also reveals that standard RL is insufficient for agentic routing without structured guidance. In ALFWorld and AppWorld, the \textit{Vanilla RL} baseline fails to identify high-leverage steps, yielding performance comparable to the \textit{Random} baseline. This failure stems from the sparsity of the reward signal, which prevents the agent from distinguishing between necessary and wasteful compute. In contrast, BoSFT successfully anchors the policy in a high-performance regime by imitating the stratified proxy. BoPO then improves upon BoSFT by optimizing the efficiency of $\mathcal{D}_{\text{easy}}$ and performance on $\mathcal{D}_{\text{hard}}$ tasks, unlocking the final gains required to dominate the efficiency frontier.

\textbf{BoPO generalizes to hard-budget constraints.} Beyond the efficiency frontier, our results in Table \ref{tab:hard_budget_sr_usepct_aligned} demonstrate that BoPO generalizes effectively to hard-budget constraints to some extent via Budget-Constrained Decoding. Moreover, in this strict setting, First-Large emerges as a remarkably strong baseline, achieving success rates that rival more complex methods. This finding underscores the critical importance of early-stage trajectory correctness: ensuring the agent is on the right track initially is often decisive for final success.
While Cascading and Single-turn routers perform similarly to this heuristic, the zero-shot GPT-5 router frequently outperforms them. This advantage likely stems from GPT-5's strong instruction-following, which allows it to interpret the remaining quota $b_t$ directly. Unlike trained baselines, it leverages this awareness to maximize high-capability calls when $K$ is large and utilize the full budget aggressively.  However, deploying such a massive model solely for routing is prohibitively expensive, underscoring the necessity for efficient, specialized routers.

\textbf{Static trade-off parameters limit adaptability to varying budget caps.}
While BoPO outperforms baselines, we acknowledge that it is not the final solution for hard-budget routing. The method relies on a fixed trade-off parameter $\lambda$ trained in a soft-budget setting, which does not perfectly generalize to rigid $K$ constraints. As shown in Table~\ref{tab:hard_budget_sr_usepct_aligned}, a small $\lambda$ is too aggressive for low budgets (leading to early exhaustion), while a large $\lambda$ is overly conservative for high budgets (failing to utilize available resources, as seen in the $K=15$ drop in AppWorld). This suggests that BoPO cannot fully adapt its risk profile dynamically based on the remaining $K$, indicating that future work requires a router explicitly trained to condition on the budget state $b_t$ to maximize utilization without violation.

\section{Analysis}
\label{sec:analysis}

\begin{figure}[t]
    % \centering
    \includegraphics[width=0.9\linewidth]{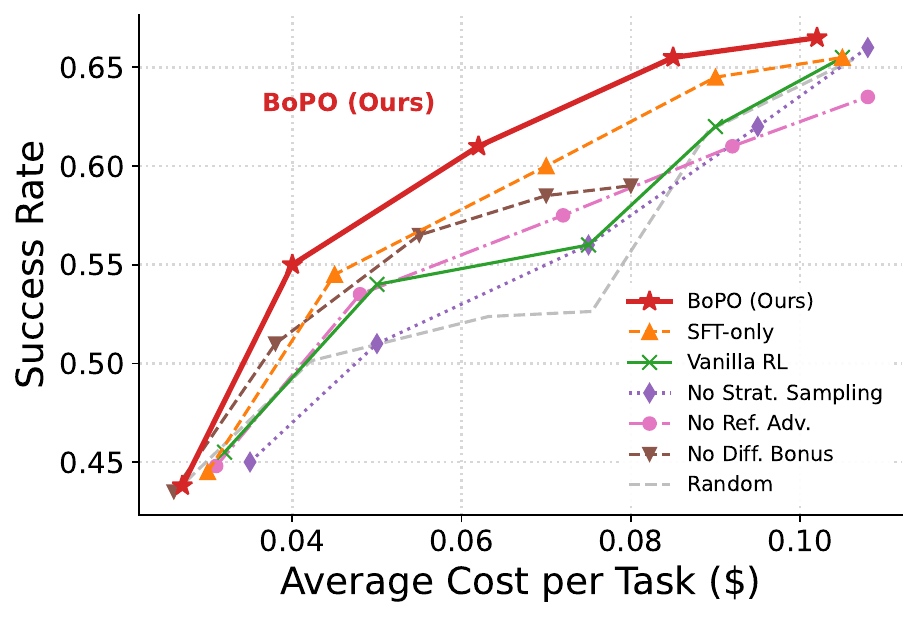}
    \caption{\textbf{Component-wise ablation study on the SciWorld benchmark.} The plot illustrates the contribution of key technical innovations to the efficiency frontier. 
    % Removing Stratified Sampling (purple) prevents the discovery of the critical path, while removing the Reference-Guided Advantage (pink) leads to instability. 
    The full BoPO method (red) strictly improves upon other ablations.}    \label{fig:ablation_study}
\end{figure}

\textbf{Ablation Study.}
To isolate the contributions of each component, we conduct a component-wise analysis on the SciWorld benchmark (Figure~\ref{fig:ablation_study}).
First, the SFT-only router outperforms the Vanilla RL baseline, confirming that our stratified sampling mechanism effectively distills an optimal path proxy that is superior to naive exploration.
In contrast, the no stratified sampling variant---which clones the raw $\pi_{large}$ trajectory for hard tasks---fails to populate the efficiency frontier, converging to a computationally wasteful policy that yields high success only at maximum cost.
Second, regarding optimization, BoPO consistently improves upon the warm-started model, whereas removing the reference-guided Advantage causes the policy to destabilize and suffer from catastrophic forgetting, effectively reverting to the performance of standard RL.
Finally, the no difficulty bonus ablation ($r_{hard}=0$) exhibits distinct performance saturation; without the explicit incentive to offset high reasoning costs, the agent adopts a ``risk-averse" strategy, prioritizing budget conservation over attempting complex tasks and resulting in premature stagnation on the efficiency curve.

\begin{figure}[t]
    \centering
    \includegraphics[width=0.85\linewidth]{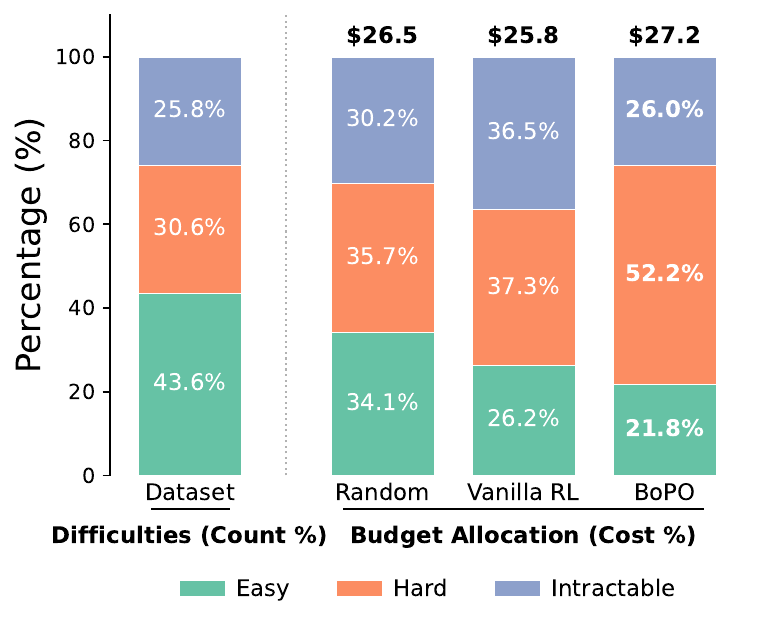}
    \caption{\textbf{Analysis of budget allocation versus task difficulty.} The leftmost bar shows the average difficulty distribution of the datasets. 
    While Random and RL baselines distribute costs inefficiently (wasting resources on Intractable or Easy tasks), BoPO (rightmost bar) strategically funds on \textit{Hard} tasks (52.2\%) where reasoning capabilities yield the highest marginal return.}    \label{fig:budget_allocation_grouped}
\end{figure}

\textbf{Budget allocation tracks where compute is most productive.}
Figure~\ref{fig:budget_allocation_grouped} contrasts the intrinsic dataset difficulty (by count) with the budget allocation (by cost) across different routers. To ensure a fair comparison, we analyze settings with similar total costs in the soft-budget mode.
We find that Random routing spreads cost more evenly across buckets, while the vanilla RL router allocates a noticeably larger share of cost to Intractable cases (37.5\%), suggesting wasted compute on instances with limited upside.
In contrast, BoPO reallocates budget away from Easy instances (21.8\%) and concentrates spending on Hard instances (52.2\%), while keeping Intractable spending closer to its prevalence (26.0\% vs.\ 25.8\% in the dataset). Meanwhile, we find taxonomy assignments are reasonably stable when varying profiling trials (Appendix \ref{app:exp}).
Overall, BoPO spends compute where it is most likely to change outcomes: less on Easy, more on Hard, and avoids over-investing in Intractable tasks.

\textbf{Generalization to Other Model Pairs.} To verify the robustness of our approach across diverse architectures, we extend our evaluation to the open-source Llama-3.1-8B-Instruct and Llama-3.1-72B-Instruct models. As illustrated in Figure~\ref{fig:llama} in the appendix, our method maintains its effectiveness, consistently optimizing the efficiency frontier with this alternative model pair.

\textbf{Latency Analysis.}
We analyze the computational overhead introduced by the router. To mitigate the significant variance inherent in direct API measurements, we report estimated end-to-end latency based on standardized token throughputs. Following OpenAI Scale Tier targets \citep{openai_scale_tier_2025}, we model generation speeds at 80 tokens/s for the large model (GPT-4.1) and 90 tokens/s for the small model (GPT-4.1 mini). For the router (Qwen2.5-1.5B), benchmarks on a standard A100 GPU using vLLM indicate a decoding speed of approximately 180 tokens/s \citep{qwen_speed_benchmark_2024}. Given that the average trajectory length across our datasets exceeds 1,000 tokens with around 20 steps, the baseline generation time is approximately 11--12 seconds per task. In contrast, the router, which generates only single-token decisions, adds a cumulative overhead of less than 0.2 seconds ($<2\%$ of total latency). This confirms that the routing cost is negligible compared to the primary generation time.

\section{Conclusion}

In this work, we formulated the problem of \textit{Budget-Aware Agentic Routing}, moving beyond single-turn model selection to address the sequential, path-dependent nature of long-horizon workflows. We introduced \textit{Boundary-Guided Training}, a novel framework that overcomes the challenges of sparse rewards by anchoring the learning process to intrinsic task difficulty. By leveraging stratified sampling to synthesize efficient critical paths and applying Boundary-Guided Policy Optimization (BoPO), our method learns to dynamically reallocate computational resources—conserving budget on easy tasks to fund complex reasoning on hard ones.
Empirically, BoPO dominates the cost-success efficiency frontier across three benchmarks, matching the performance of expensive baselines at significantly reduced costs. While our Budget-Constrained Decoding mechanism effectively enforces hard limits at inference time, our analysis highlights that static reward parameters do not fully adapt to varying budget caps. Future work should therefore focus on training routers that explicitly condition on the residual budget state to maximize utilization without violation.

\section*{Impact Statement}

This paper introduces \textit{Budget-Aware Agentic Routing} to improve the economic viability of autonomous agents. Our work primarily advances machine learning efficiency, but has broader implications for accessibility and sustainability.

\textbf{Broader Impact.} By matching the performance of large models at a fraction of the cost, our method lowers the barrier to entry for advanced AI, enabling resource-constrained organizations to utilize complex agentic workflows. Furthermore, by routing the majority of steps to smaller, less energy-intensive models, we can reduce the aggregate energy consumption and carbon footprint of agentic deployments compared to standard practices. We do not foresee immediate negative societal consequences that must be specifically highlighted here.

\bibliography{example_paper}
\bibliographystyle{icml2026}

\begin{table*}[h!]
\centering
\caption{Reward Landscape across Task Taxonomies. The objective function ensures that success is always preferable to failure, while the magnitude of the positive signal dictates the allowable budget for each task type. *Note that success on Intractable tasks is rare by definition; the reward structure defaults the agent to a cost-minimizing behavior similar to Easy tasks.}

\label{tab:reward_landscape}
\resizebox{\textwidth}{!}{%
\begin{tabular}{@{}lllll@{}}
\toprule
\textbf{Task Type} & \textbf{Outcome} & \textbf{Reward Components} & \textbf{Total Reward} $R(\tau, x)$ & \textbf{Strategic Implication} \\ \midrule
\multirow{2}{*}{\textbf{Easy}} & Success & $r_{\text{success}}$ & $r_{\text{success}} - \lambda \cdot \mathcal{C}_{\text{norm}}$ & \textbf{Maximize Efficiency:} Margin is thin; high cost erases profit. \\
 & Failure & $0$ & $-\lambda \cdot \mathcal{C}_{\text{norm}}$ & \textbf{Avoid:} Failure is strictly worse than cheap success. \\ \midrule
\multirow{2}{*}{\textbf{Hard}} & Success & $r_{\text{success}} + r_{\text{hard}}$ & $(r_{\text{success}} + r_{\text{hard}}) - \lambda \cdot \mathcal{C}_{\text{norm}}$ & \textbf{Prioritize Solving:} High bonus justifies expensive compute. \\
 & Failure & $0$ & $-\lambda \cdot \mathcal{C}_{\text{norm}}$ & \textbf{Avoid:} Expensive failure is the worst-case scenario. \\ \midrule
\multirow{2}{*}{\textbf{Intractable}} & Success$^*$ & $r_{\text{success}}$ & $r_{\text{success}} - \lambda \cdot \mathcal{C}_{\text{norm}}$ & \textbf{Opportunistic:} Treated as a standard win if solved by chance. \\
 & Failure & $0$ & $-\lambda \cdot \mathcal{C}_{\text{norm}}$ & \textbf{Fail Cheaply:} No bonus incentives exist; minimize loss. \\ \bottomrule
\end{tabular}%
}
\vspace{0.2cm}
\end{table*}

\newpage
\appendix

\section{Limitations and Future Work}

\textbf{Training Overhead.} To overcome the challenge of sparse rewards in long-horizon tasks, our Boundary-Guided Training pipeline requires constructing a difficulty taxonomy via profiling and synthesizing data through stratified sampling. While this incurs a higher upfront computational cost compared to standard RL or single-turn preference optimization, this is a one-time fixed cost. In practical deployments, this expense is amortized over millions of inference runs, where the resulting efficiency gains from reduced large-model usage provide significant long-term savings.

\textbf{Model Scope.} We intentionally focused this study on the binary choice between a cost-efficient model ($\mathcal{M}_{small}$) and a high-capability model ($\mathcal{M}_{large}$) to isolate the fundamental economic trade-off. However, our mathematical formulation is model-agnostic and easily extensible. Generalizing to $N$ models simply requires expanding the action space to $\mathcal{A} = \{\mathcal{M}_1, \dots, \mathcal{M}_N\}$ and defining the associated cost function $c(a_t)$ in Eq.~(1), without altering the fundamental reward structure or optimization procedure.

\textbf{Static Risk Profiles.} Our method trains the router to optimize a soft-budget efficiency frontier using a fixed trade-off parameter $\lambda$. Although we provide the residual budget $b_t$ in the system prompt and enforce limits via Budget-Constrained Decoding at inference, the policy itself is not \textit{trained} to dynamically adjust its risk appetite based on $b_t$. Consequently, the learned behavior reflects a static risk profile defined by $\lambda$, which may be suboptimal—either too aggressive or too conservative—when transferred to varying hard budget caps ($K$). Future work should bridge this gap by explicitly conditioning the policy optimization on the budget state to interpolate smoothly between behaviors.

\section{Reward Landscape Analysis}
\label{app:reward_analysis}

To provide a closer look at how the Boundary-Relative objective governs agent behavior, we detail the reward outcomes across all task difficulty profiles in Table~\ref{tab:reward_landscape}.

The formulation creates three distinct behavioral incentives:
\begin{enumerate}
    \item \textbf{Frugality on Easy Tasks:} Since the base reward $r_{\text{success}}$ is small, the only way to achieve a high net reward is to keep the cost term $-\lambda \cdot \mathcal{C}_{\text{norm}}$ near zero.
    \item \textbf{Investment in Hard Tasks:} The difficulty premium $r_{\text{hard}}$ effectively subsidizes the cost of reasoning, allowing the agent to ``spend" budget without penalizing the net objective.
    \item \textbf{Risk Aversion on Intractable Tasks:} Because Intractable tasks do not trigger the difficulty premium (and success is statistically unlikely), the agent defaults to the logic of Easy tasks: minimizing cost to mitigate the inevitable penalty of failure.
\end{enumerate}

\section{Challenges of Direct Hard-Budget Training}
\label{app:direct_hard_training}

While our method trains on the soft efficiency frontier in Eq.~\eqref{eq:soft} and enforces strict feasibility at inference time via Budget-Constrained Decoding (BCD), a natural alternative would be to train the routing policy to satisfy the hard constraint in Eq.~\eqref{eq:hard} end-to-end. Here, we formalize why solving Eq.~\eqref{eq:hard} directly is theoretically and practically challenging for long-horizon agentic workflows with sparse terminal feedback.

Recall that at each step $t$, the router observes the interaction history $s_t$ and chooses $a_t \in \mathcal{A}=\{\mathcal{M}_{\text{small}},\mathcal{M}_{\text{large}}\}$ according to $\pi_\theta(a_t \mid s_t)$, incurring cost $c(a_t)>0$ and inducing transitions $s_{t+1}\sim P(\cdot\mid s_t,a_t)$. A trajectory is $\tau=\{(s_t,a_t)\}_{t=0}^{|\tau|-1}$ with sparse terminal indicator $\mathbb{I}(\mathrm{success}(\tau))\in\{0,1\}$. Define the total trajectory cost as
\begin{equation}
C(\tau)\;=\;\sum_{t=0}^{|\tau|-1} c(a_t).
\end{equation}
The hard-budget CMDP in Eq.~\eqref{eq:hard} seeks to maximize success subject to the path-wise constraint $C(\tau)\le B_{\max}$.

Compared to the soft-penalty objective in Eq.~\eqref{eq:soft}, directly optimizing Eq.~\eqref{eq:hard} differs in three critical ways:

\paragraph{(1) Combinatorial, non-smooth feasible action sets.}
The hard-budget constraint in Eq.~\eqref{eq:hard} is a \emph{sequence-level} constraint: whether a trajectory is allowed depends on the entire routing sequence, not on marginal action frequencies. In the common case where the budget effectively counts expensive calls (e.g., after normalization $c(\mathcal{M}_{\text{small}})=0$ and $c(\mathcal{M}_{\text{large}})=1$), the constraint $C(\tau)\le B_{\max}$ becomes
\[
\#\{t: a_t=\mathcal{M}_{\text{large}}\}\le K,\qquad K=\lfloor B_{\max}\rfloor.
\]
For a fixed horizon $T$, the number of feasible length-$T$ routing sequences is exactly
\begin{equation}
|\mathcal{A}_{\mathrm{feas}}(T,K)| \;=\; \sum_{k=0}^{K} \binom{T}{k}.
\end{equation}
Therefore the feasible set is \emph{combinatorial} and can change abruptly with $B_{\max}$ (or the realized length $|\tau|$). Learning is harder because the policy must decide \emph{when} to spend scarce $\mathcal{M}_{\text{large}}$ calls, not just \emph{how many} to use on average.

\paragraph{(2) Discontinuous objectives induce boundary-dominated, high-variance gradients.}
A standard way to enforce hard feasibility during training is to reject, truncate, or mask trajectories/actions that violate the constraint. This yields a discontinuous constrained return, e.g.,
\begin{equation}
J_{\mathrm{hard}}(\theta)
\;=\;
\mathbb{E}_{\tau\sim \pi_\theta}\!\left[
\mathbb{I}(\mathrm{success}(\tau)) \cdot \mathbf{1}\{C(\tau)\le B_{\max}\}
\right].
\label{eq:appendix_Jhard_indicator}
\end{equation}
In this form, the learning signal is nonzero only on trajectories that are \emph{both successful and feasible}. In long-horizon agentic tasks, $\mathbb{I}(\mathrm{success}(\tau))$ is sparse, and the additional feasibility event $\{C(\tau)\le B_{\max}\}$ further shrinks the set of informative samples. Consequently, Monte Carlo policy-gradient estimators become ill-conditioned: their variance increases sharply as the probability of sampling successful-feasible trajectories decreases. Moreover, feasibility changes discontinuously at the boundary $C(\tau)=B_{\max}$, so trajectories just below/above the boundary contribute radically different returns; gradient estimates then become dominated by rare ``near-boundary'' events, leading to unstable optimization or near-zero updates.

\paragraph{(3) Resource-state dependence and conservative collapse in sparse-reward CMDPs.}
Optimal hard-budget routing is inherently budget-dependent: the best action at time $t$ depends on the remaining budget
\begin{equation}
b_t \;=\; B_{\max} - \sum_{i=0}^{t-1} c(a_i),
\end{equation}
so the effective decision state is $(s_t,b_t)$ (even if $s_t$ already encodes the interaction history). Without explicit conditioning on $b_t$ and sufficient exploration of low-budget ``boundary'' regimes, the learner faces a severe exploration--exploitation dilemma: it must discover when $\mathcal{M}_{\text{large}}$ is worth spending despite long-horizon credit assignment and sparse terminal feedback. In practice, common CMDP solvers rely on Lagrangian relaxation,
\begin{equation}
\min_{\lambda\ge 0}\ \max_{\theta}\ 
\mathbb{E}_{\tau\sim \pi_\theta}\!\left[
\mathbb{I}(\mathrm{success}(\tau)) - \lambda\big(C(\tau)-B_{\max}\big)
\right],
\label{eq:appendix_lagrangian}
\end{equation}
which introduces a minimax game. In sparse-reward regimes, early dual updates can increase $\lambda$ before the policy learns to succeed, making the locally easiest way to reduce violations to adopt a conservative strategy (e.g., near ``always-$\mathcal{M}_{\text{small}}$''). This ``safe'' collapse then suppresses sampling of trajectories that require $\mathcal{M}_{\text{large}}$ to achieve $\mathbb{I}(\mathrm{success}(\tau))=1$, eliminating the very data needed to learn budget-efficient success on difficult tasks.

\smallskip
Taken together, hard-budget training couples (i) combinatorial sequence-level feasibility, (ii) discontinuous, boundary-driven learning signals, and (iii) resource-dependent long-horizon credit assignment under sparse reward. These properties make direct end-to-end optimization of Eq.~\eqref{eq:hard} substantially less stable and sample-efficient than training with the smooth soft objective in Eq.~\eqref{eq:soft} and enforcing strict feasibility at inference time.

\section{Experiment Details}
\label{app:exp}

\textbf{Success Rate Calculation.} We align our success metrics with the standard evaluation protocols for each benchmark. For \sci, which assigns a granular progress score (0--100), we report the average episode score. For \alf, we utilize the standard binary success indicator (0 or 1). For \app, we report the Task Goal Completion (TGC) score from original evaluation script.

\textbf{Pipeline details.} (i) \textbf{Difficulty profiling:} for each task $x$, we execute $\pi_{\text{small}}$ and $\pi_{\text{large}}$ for $K{=}5$ trials and partition tasks into $\mathcal{D}_{\text{easy}}$, $\mathcal{D}_{\text{hard}}$, and $\mathcal{D}_{\text{intractable}}$. (ii) \textbf{Boundary-Guided SFT:} for Hard tasks, we generate $N{=}20$ stratified trajectories by sweeping $p_k=k/N$ and select the minimum-cost successful trajectory as the expert; Easy/Intractable tasks use $\pi_{\text{small}}$ trajectories. We fine-tune the router to emit the decision tokens \textsc{SMALL}/\textsc{LARGE}, masking loss to the decision positions, and oversample Hard tasks to comprise $\approx 70\%$ of each batch. (iii) \textbf{BoPO:} starting from the SFT router, we optimize the boundary-relative reward with the reference-guided advantage and a KL penalty to the SFT reference. (iv) \textbf{Hard budgets:} at inference, we apply Budget-Constrained Decoding by injecting remaining budget $b_t$ into the router prompt and pruning $M_{\text{large}}$ when it would violate $B_{\max}$.

\textbf{Balanced Training.}
We fine-tune the router to predict a restricted action space mapped to two special tokens (\texttt{SMALL}, \texttt{LARGE}). We employ standard cross-entropy loss, applying masks to ensure gradients are backpropagated solely from these decision tokens. To counteract the ``laziness" induced by the natural prevalence of Easy tasks, we curate a balanced batch distribution where Hard tasks are oversampled ($\approx 70\%$). This configuration forces the model to prioritize learning the complex decision boundaries required for high-reasoning tasks, rather than collapsing into a trivial ``always-small" policy.

\begin{table}[h]
\centering
\caption{Dataset statistics and constraints used in our experiments.}
\label{tab:datasets}
\begin{small}
\begin{sc}
\begin{tabular}{lccc}
\toprule
Dataset & Train Size & Test Size & Max Steps \\
\midrule
\alf & 2420 & 200 & 30 \\
\sci & 2120 & 200 & 40 \\
\app & 105 & 168 & 40 \\
\bottomrule
\end{tabular}
\end{sc}
\end{small}
\end{table}

\paragraph{Reproducibility and implementation details.}
To ensure reproducibility, we provide a complete specification of the training and evaluation pipeline: (i) the exact router input formatting and serialization of $s_t$ (truncated to the most recent 2k tokens), the decision-token masking scheme, and the full prompt templates; (ii) all optimization hyperparameters (global batch size 64; BoSFT learning rate $2\times 10^{-5}$ with cosine decay and 3\% warmup; BoPO learning rate $1\times 10^{-6}$); and (iii) GRPO/BoPO configuration (group size $G=8$, KL coefficient $\beta=0.04$, and reward weights $r_{\text{success}}=1.0$, $r_{\text{hard}}=0.5$, $\lambda \in \{0.1, 0.3, 0.5, 0.7, 0.9\}$). 
Episodes terminate at 30 steps (ALFWorld) and 40 steps (SciWorld/AppWorld). For cost reporting we use OpenAI list prices (as of submission), corresponding to a large-to-small price ratio of 5$\times$; we also report large-call counts to ensure price-agnostic comparison. Codes will be released upon acceptance.

\paragraph{Robustness of the difficulty taxonomy to profiling budget.}
To test whether the difficulty taxonomy (\S4.1) depends on the profiling trial count, we repeat profiling with $K\in\{3,10\}$. Relative to the default $K=5$, the per-task label agreement is 78.5\% for $K=3$ and 87.4\% for $K=10$; for the critical $\mathcal{D}_{\text{hard}}$ subset, the Jaccard similarity is 0.69 for $K=3$ and 0.76 for $K=10$. This indicates that reasoning-intensive tasks are consistently identified despite stochasticity. Notice that taxonomy is solely for constructing training signals; the router never sees labels at test time.

\section{Computational Cost of the Router}
\label{app:cost_router}

In the efficiency frontiers presented in Figure \ref{fig:pareto_frontiers} and the hard-budget evaluations in Table \ref{tab:hard_budget_sr_usepct_aligned}, we excluded the computational cost of the routing module itself from the reported metrics.

For our trained router and similar baselines (e.g., Single-turn, Cascading), this exclusion is justified by their minimal overhead. As detailed in the Latency Analysis (Section \ref{sec:analysis}), these routers utilize lightweight architectures (e.g., Qwen2.5-1.5B) to generate a single decision token per step. This process incurs a latency overhead of less than 0.2 seconds—representing under 2\% of the total trajectory execution time—making the monetary cost negligible compared to the agentic workflow itself.

In contrast, the Zero-shot GPT-5 router is included primarily as a performance upper bound rather than a practical deployment candidate. While it demonstrates strong instruction-following capabilities that are advantageous for strict hard-budget constraints, it often lacks the granularity to adapt dynamically in soft-budget settings, frequently collapsing to static policies. Overall, utilizing such a massive model solely for routing is economically self-defeating; our results demonstrate that specialized, lightweight routers like BoPO are essential to achieve comparable decision-making quality without consuming the very budget the system aims to save.

\begin{figure}[t]
    \centering
    \includegraphics[width=0.99\linewidth]{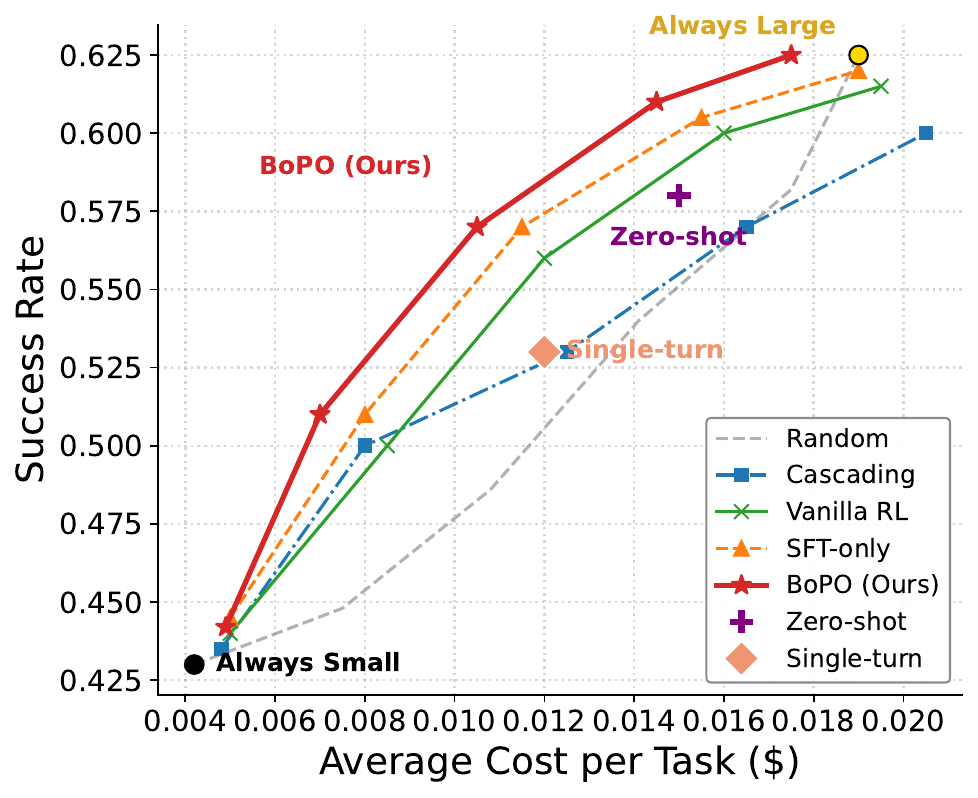}
    \caption{\textbf{Generalization to open-source model pairs.} Pareto efficiency frontier on the SciWorld benchmark using Llama-3.1-8B-Instruct as $\mathcal{M}_{small}$ and Llama-3.1-72B-Instruct as $\mathcal{M}_{large}$. BoPO (red curve) consistently outperforms baselines, demonstrating that our boundary-guided framework is model-agnostic and successfully transfers to architectures with different cost-capability ratios. We use the API provided by \textit{together.ai}. }    \label{fig:llama}
\end{figure}

\section{Analysis of Cost Inversions.}

\begin{figure*}[h]
    \centering
    \includegraphics[width=0.98\linewidth]{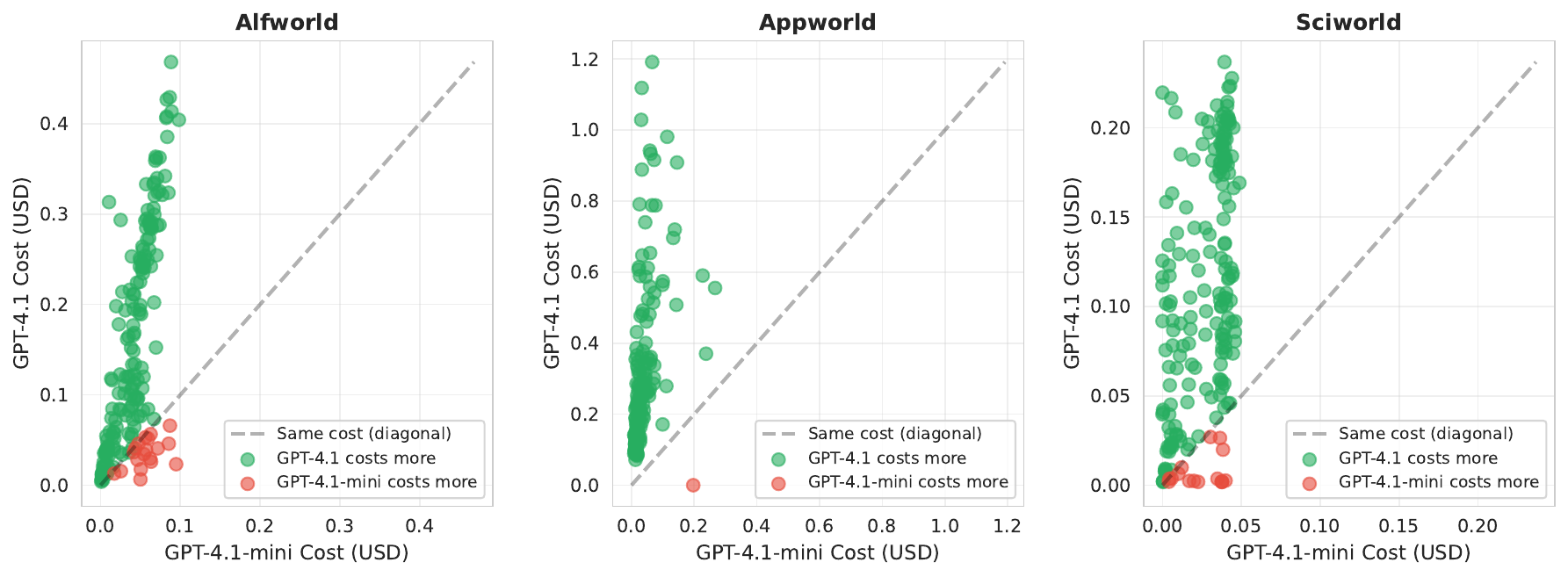}
    \caption{\textbf{Cost distribution analysis between static policies.} We plot the trajectory cost of $\pi_{large}$ (y-axis) versus $\pi_{small}$ (x-axis) across three benchmarks. Green points ($\mathcal{C}_{large} > \mathcal{C}_{small}$) represent standard cases, while red points indicate \textbf{cost inversions} where the small model incurs higher costs due to long, ineffective failure trajectories. This high variance and frequent inversion necessitate our \textbf{Normalized Cost} formulation, which scales penalties relative to the specific task's difficulty boundaries rather than absolute dollar values.}    \label{fig:cost_comparison}
\end{figure*}

As illustrated in Figure~\ref{fig:cost_comparison}, a counter-intuitive phenomenon occurs in agentic workflows where ``cheaper'' models become more expensive than ``expensive'' models (red data points). This typically happens when the weaker model ($\mathcal{M}_{small}$) fails to solve the task efficiently, entering repetitive loops or generating long, incorrect chains of thought that consume the maximum context window. If the reward function relied on raw scalar cost, the agent would receive a disproportionately high penalty for these ``cheap failures,'' potentially destabilizing training. Our \textbf{Normalized Cost} $\mathcal{C}_{norm}(\tau, x)$ addresses this by clipping and scaling the cost relative to the specific task's $C_{min}$ and $C_{max}$ boundaries. This ensures that the penalty reflects the \textit{relative} efficiency of the trajectory within the task's intrinsic difficulty, rather than the absolute token usage, making the optimization robust to the high variance seen in Figure~\ref{fig:cost_comparison}.

\section{Prompts}
\label{app:prompts}

\begin{table*}[h!]
\centering
\footnotesize
\caption{\textbf{Soft Budget Routing Mode Prompt.} The router uses a sliding window of history to make a binary decision.}
\label{tab:standard-routing-prompt}
\begin{tabular}{p{0.95\textwidth}}
\toprule
\textbf{System Prompt Configuration} \\
\midrule
\prompttext{You are a routing agent that decides which language model to use for each step of a task.} \\
\\
\prompttext{Task: \promptvar{\{task\_desc\}}} \\
\\
\prompttext{You must choose between:} \\
\prompttext{- SMALL: A fast, efficient model (GPT-4.1-mini) suitable for simple tasks} \\
\prompttext{- BIG: A powerful model (GPT-4.1) for complex reasoning} \\
\\
\prompttext{Previous Steps:} \\
\metainstruct{[If no history]} \\
\prompttext{No previous steps.} \\
\\
\metainstruct{[If history exists, for each step i in recent history]} \\
\prompttext{Step \promptvar{\{i\}} [Model: \promptvar{\{model\}}]:} \\
\prompttext{\quad Action: \promptvar{\{action\_preview\}}} \\
\prompttext{\quad Result: \promptvar{\{observation\_preview\}}} \\
\\
\prompttext{For the next step, which model should be used? Respond with either SMALL or BIG.} \\
\prompttext{Decision:} \\
\bottomrule
\end{tabular}
\end{table*}

\begin{table*}[h!]
\centering
\footnotesize
\caption{\textbf{Hard Budget Mode Prompt.} Includes explicit budget state tracking and conditional fallback logic.}
\label{tab:hard-budget-prompt}
\begin{tabular}{p{0.95\textwidth}}
\toprule
\textbf{System Prompt Configuration} \\
\midrule
\prompttext{You are a budget-aware routing agent that decides which language model to use for each step.} \\
\\
\prompttext{Task: \promptvar{\{task\_desc\}}} \\
\prompttext{Environment: \promptvar{\{context\}}} \\
\\
\prompttext{BUDGET CONSTRAINTS:} \\
\prompttext{- Maximum BIG model calls allowed: \promptvar{\{max\_big\_budget\}}} \\
\prompttext{- BIG calls used so far: \promptvar{\{big\_used\}}} \\
\prompttext{- BIG calls remaining: \promptvar{\{big\_left\}}} \\
\prompttext{- Current step: \promptvar{\{current\_step\}} / \promptvar{\{max\_steps\}}} \\
\\
\prompttext{You must choose between:} \\
\prompttext{- SMALL: A fast, efficient model (GPT-4.1-mini) suitable for simple tasks} \\
\prompttext{- BIG: A powerful model (GPT-4.1) for complex reasoning and difficult tasks} \\
\\
\metainstruct{[If budget remaining ($big\_left > 0$)]} \\
\prompttext{HARD BUDGET MODE:} \\
\prompttext{You have a strict budget of \promptvar{\{max\_big\_budget\}} BIG calls. Once exhausted, you can ONLY use SMALL.} \\
\prompttext{Plan carefully: with \promptvar{\{big\_left\}} BIG call(s) remaining and \promptvar{\{max\_steps - current\_step + 1\}} steps left,} \\
\prompttext{consider when BIG will be most valuable (e.g., complex reasoning, getting unstuck, critical decisions).} \\
\\
\metainstruct{[If budget exhausted ($big\_left \le 0$)]} \\
\prompttext{HARD BUDGET MODE - BUDGET EXHAUSTED:} \\
\prompttext{You have used all \promptvar{\{max\_big\_budget\}} allowed BIG calls. You MUST use SMALL for remaining steps.} \\
\prompttext{Focus on completing the task efficiently with the SMALL model only.} \\
\\
\prompttext{Previous Steps:} \\
\metainstruct{[If history exists]} \\
\prompttext{(Steps 1-\promptvar{\{num\_omitted\}} omitted)} \metainstruct{[if applicable]} \\
\\
\metainstruct{[For each step in recent history (last 10 steps)]} \\
\prompttext{Step \promptvar{\{actual\_step\_num\}} [Model: \promptvar{\{model\}}]:} \\
\prompttext{\quad Action: \promptvar{\{action\_preview\}}} \\
\prompttext{\quad Result: \promptvar{\{observation\_preview\}}} \\
\\
\metainstruct{[If budget exhausted]} \\
\prompttext{Since your budget is exhausted, you must respond with: SMALL} \\
\prompttext{Decision:} \\
\\
\metainstruct{[If budget remaining]} \\
\prompttext{For the next step, which model should be used? Respond with either SMALL or BIG.} \\
\prompttext{Consider the budget constraints and remaining steps carefully.} \\
\prompttext{Decision:} \\
\bottomrule
\end{tabular}
\end{table*}

\end{document}